\documentclass{article}

 \usepackage[preprint]{neurips_2026}

\usepackage[utf8]{inputenc} 
\usepackage[T1]{fontenc}    
\usepackage{hyperref}       
\usepackage{url}            
\usepackage{booktabs}       
\usepackage{amsfonts}       
\usepackage{nicefrac}       
\usepackage{microtype}      
\usepackage{xcolor}         
\usepackage{graphicx}
\usepackage{threeparttable}
\usepackage{booktabs}
\usepackage{amsmath}
\usepackage{amssymb}
\usepackage{graphicx}
\usepackage{multirow}
\usepackage{array}
\usepackage{enumitem}
\usepackage{xspace}  
\usepackage{amsthm}   
\theoremstyle{remark} 
\usepackage{threeparttable} 
\usepackage{subcaption} 

\newcommand{\method}{I\textsuperscript{2}RiMA}
\newcommand{\R}{\mathbb{R}}
\newcommand{\SPD}{\mathcal{S}_{++}^{C}}

\title{I\textsuperscript{2}RiMA: Spectral Riemannian Representation with Temporal Attention for Mental Stress Detection based on EEG Signals}
\author{
  Cheng He$^{*1}$ \And
  Kunyu Peng$^{*2}$ \And
  Shangen Han$^{*1}$ \And
  Jinming Ma$^{1}$ \And
  Jinhong Ding$^{3}$ \And
  Likun Xia$^{1\dagger}$ \\[2mm]
  $^{1}$ Laboratory of Neural Computing and Intelligent Perception, \\
  College of Information Engineering, Capital Normal University, Beijing, China \\
  $^{2}$ Karlsruhe Institute of Technology, Germany \\
  $^{3}$ School of Psychology, Capital Normal University, Beijing, China
}
\begin{document}
\maketitle
\begin{abstract}
Cross-subject EEG stress detection remains challenging because discriminative stress-related patterns are both subject-dependent and frequency-specific. Conventional Riemannian methods model spatial covariance mainly in the time domain, overlooking neural oscillations that are critical for high-level cognitive state decoding, while standard temporal tokenization often fragments inter-slice temporal coherence. To address these limitations, we propose \method{}, an Intra-Inter Riemannian Manifold Attention Network for EEG-based stress detection. \method{} constructs spatial covariance matrices independently at each frequency point and maps them to the SPD tangent space, preserving channel-wise geometry together with frequency-specific discriminative cues. It further introduces frequency cluster aggregation to select informative spectral components and reduce redundancy by forming compact, data-driven frequency clusters aligned with EEG rhythms. Finally, an intra-inter slice attention module adaptively integrates local slice-level spectral dynamics and global temporal context across EEG sequences. Experiments on three datasets show that \method{} consistently outperforms five state-of-the-art baselines, achieving up to 82.78\% balanced accuracy while remaining efficient with only 1.60M parameters and 31.95M FLOPs.
\end{abstract}

\section{Introduction}

Mental stress is a widespread public health issue, especially among young people, with over 60\% reporting chronic moderate-to-severe stress~\cite{fu2023report, steel2014global}. As prolonged stress increases the risk of depression, anxiety, and cardiovascular disease~\cite{meier2015secondary}, objective and scalable stress detection is urgently needed. EEG offers a promising solution due to its millisecond temporal resolution, non-invasive and portable nature, and sensitivity to psychological states~\cite{da2013eeg}. Recent wireless dry-electrode systems further support everyday EEG-based stress monitoring beyond clinical settings~\cite{arpaia2020wearable}.

Considerable work has explored EEG-based stress detection using traditional machine learning, deep learning, and Riemannian geometry. However, existing methods face two key limitations. First, conventional Riemannian approaches usually build covariance matrices from time-domain EEG signals~\cite{mognon2011adjust}, which is suitable for low-level tasks with temporally localized responses, such as visual evoked potential decoding~\cite{wang2024grassmannian}. In contrast, high-level states like stress and emotion involve distributed, spectrally structured neural dynamics, making time-domain channel covariance insufficient. Although frequency-domain covariance is a natural alternative, naive construction can mix frequency information and obscure physiologically meaningful band-specific patterns. Moreover, EEG covariance matrices lie on the SPD manifold~\cite{barachant2010riemannian}; direct Euclidean vectorization ignores this geometry and causes information loss, especially under cross-subject distribution shifts~\cite{yger2016riemannian}.

Second, most existing EEG tokenization pipelines segment continuous signals into fixed-length windows and process each window independently. This design discards temporal coherence across adjacent slices, even though such context has been shown to carry discriminative information for mental state classification~\cite{grissmann2017context}. Recent transformer-based EEG models~\cite{yang2023biot, jiang2024large, li2024neurobolt} have improved sequence modeling, but they still often treat each EEG slice as an isolated token. As a result, they struggle to jointly capture intra-slice local spectral structure and inter-slice global temporal dependencies, both of which are essential for robust high-level cognitive state recognition.

Motivated by these observations, we propose \method{}, an Intra-Inter Riemannian Manifold Attention Network for EEG-based stress detection. \method{} is built on two key insights. First, stress- and emotion-related EEG signals exhibit state-dependent, frequency-specific spatial patterns. Since distinct EEG bands reflect different neural dynamics~\cite{buzsaki2012origin}, \method{} constructs covariance matrices independently at each frequency point and maps them to the SPD tangent space, preserving channel-wise Riemannian geometry without prematurely mixing spectral information. A frequency cluster aggregation module further selects informative frequency-specific Riemannian features and reduces redundancy, yielding compact and interpretable representations.
Second, longer temporal windows improve intra-slice separability while reducing inter-slice variability, suggesting complementary local and global temporal cues. To capture this structure, \method{} introduces an intra-inter slice attention fusion module that adaptively aggregates EEG slices via linear encoding, attention weighting, and weighted fusion, preserving local discriminative patterns while modeling sequence-level temporal context.
Together, the frequency-aware Riemannian representation and intra-inter slice attention fusion enable geometry-preserving, physiologically grounded, and temporally contextualized EEG representations for robust cross-subject stress detection.

Our main contributions are summarized as follows:
\begin{enumerate}[leftmargin=*, itemsep=1pt]
\item We propose \method{}, a frequency-aware Riemannian network for cross-subject EEG stress detection. It constructs covariance matrices at each frequency point and maps them to the SPD tangent space, preserving channel-wise geometry and frequency-specific discriminative patterns.

\item We introduce frequency cluster aggregation for data-driven feature selection and redundancy reduction, together with intra-inter slice attention fusion to capture both local slice-level patterns and global temporal dependencies in continuous EEG.

\item Experiments show that \method{} achieves state-of-the-art B.ACC of 77.59\%, 75.88\%, and 82.78\% on MIST Control, MIST Stress, and SEED, respectively, while using only 1.60M parameters and 31.95M FLOPs.


\end{enumerate}

\section{Related Work}

\noindent\textbf{EEG-Based Stress Detection}
EEG-based stress detection methods generally follow two paradigms: traditional machine learning and deep learning. Traditional methods rely on handcrafted temporal, spectral, or nonlinear features with classifiers such as SVM and KNN. Representative studies include feature fusion for workload identification~\cite{pei2020eeg}, frontal alpha asymmetry for real-time stress assessment~\cite{arpaia2020wearable}, multi-level stress classification with combined EEG and ECG features~\cite{xia2018a}, and linear/nonlinear EEG features for depression detection~\cite{cai2018study}. Deep learning methods enable end-to-end representation learning by capturing spatial, spectral, and temporal EEG patterns, including spatio-temporal modeling for cross-subject motor imagery decoding~\cite{lv2025enhanced}, R3DCNN for workload assessment~\cite{zhang2018learning}, MuLHiTA with multi-branch LSTM and hierarchical attention~\cite{xia2022mulhita}, and phase-space reconstruction with geometric features for depression detection~\cite{akbari2021depression}. Despite their progress, many methods remain limited to single-subject or small-scale evaluations and often suffer substantial performance degradation in cross-subject settings due to inter-subject variability~\cite{giannakakis2019review}.

\noindent\textbf{Cross-Subject EEG Learning.}
Cross-subject generalization is a core challenge in EEG decoding. BIOT~\cite{yang2023biot} employs channel embedding alignment to mitigate individual differences, LaBraM~\cite{jiang2024large} leverages large-scale pre-training for enhanced generalization, and NeuroBOLT~\cite{li2024neurobolt} synthesizes fMRI signals from raw EEG via multi-dimensional representation learning. Domain adaptation and transfer learning are dominant strategies~\cite{lotte2018review}, yet they primarily address cross-subject distribution shifts while neglecting intra-subject temporal domain drift (ISTDS)~\cite{jayaram2016transfer}---the phenomenon where a single subject's EEG distribution shifts across experimental sessions due to fatigue, electrode impedance changes, and physiological fluctuations~\cite{zanetti2021multilevel}.

\noindent\textbf{Riemannian Learning for EEG.}
Covariance matrices of multi-channel EEG signals lie on the SPD manifold $\SPD$, which possesses non-Euclidean geometric structure. Traditional vectorization in Euclidean space destroys this structure, leading to information loss~\cite{barachant2010riemannian}. The affine-invariant Riemannian metric (AIRM) provides robustness to individual differences~\cite{barachant2011multiclass}, while the Log-Euclidean metric offers computational efficiency with geometric consistency~\cite{congedo2017riemannian}. Riemannian methods have demonstrated significant performance improvements in BCI and EEG classification, particularly in cross-subject scenarios~\cite{yger2016riemannian}. However, existing Riemannian approaches typically construct a single covariance matrix from time-domain signals, mixing frequency information and obscuring frequency-specific spatial patterns~\cite{congedo2017riemannian}.

\noindent\textbf{Temporal Aggregation and Attention Modeling.}
Transformer-based methods~\cite{yang2023biot, jiang2024large, li2024neurobolt} model EEG sequences at the token level, but often process each slice independently, overlooking both intra-slice spectral dynamics and inter-slice temporal coherence. Although hierarchical and manifold attention mechanisms~\cite{xia2022mulhita, pan2022matt} have been explored for temporal aggregation, they typically assume stable feature distributions across subjects and sessions. In contrast, \method{} jointly integrates frequency-aware Riemannian manifold learning with intra-inter slice attention fusion for cross-subject EEG stress detection, addressing this gap~\cite{giannakakis2019review}.

\section{Problem Formulation}

\paragraph{Input Definition}
Let $\mathcal{D} = \{(\mathbf{X}_u, y_u)\}_{u=1}^{U}$ denote an EEG dataset where $\mathbf{X}_u \in \R^{C \times L}$ is the $u$-th trial with $C$ channels and $L$ samples, and $y_u \in \{1, \ldots, G\}$ is the class label. Each trial is segmented into $m$ non-overlapping slices $\mathcal{S} = \{s_1, \ldots, s_m\}$ with $s_i \in \R^{C \times T}$. The FFT transforms each slice into the frequency domain: $\mathbf{X}_{\text{freq}} \in \R^{B \times m \times C \times F}$, with $B$ batch size and $F$ frequency points.

\paragraph{Task Definition}
We define the cross-subject stress detection task as learning a classifier $f_\theta$ that generalizes to unseen subjects. Formally, given training subjects $\mathcal{P}_{\text{train}}$ and test subjects $\mathcal{P}_{\text{test}}$ with $\mathcal{P}_{\text{train}} \cap \mathcal{P}_{\text{test}} = \emptyset$, the objective is to minimize the expected cross-entropy loss $\ell$ over the test distribution, as defined in Eq.~\ref{eq:task_obj}:
\begin{equation}
    \theta^* = \arg\min_\theta \; \mathbb{E}_{(\mathbf{X}, y) \sim \mathcal{D}_{\text{test}}} \left[ \ell\left(f_\theta(\mathbf{X}), y\right) \right]
    \label{eq:task_obj}
\end{equation}

\paragraph{Learning Objective}
We seek a representation mapping $\phi: \R^{C \times L} \to \R^d$ that satisfies three properties: (1)~\textit{geometry preservation}---$\phi$ retains the spatial structure of EEG signals via Riemannian manifold modeling; (2)~\textit{temporal coherence}---$\phi$ encodes both intra-slice local spectral patterns and inter-slice global dependencies through attention fusion; (3)~\textit{cross-subject robustness}---$\phi$ is invariant to inter-subject distribution shifts.

\section{Method}

\subsection{\texorpdfstring{Overview of \method{}}{Overview of I2RiMA}}

\begin{figure}[t]
    \centering
    \includegraphics[width=1.0\linewidth]{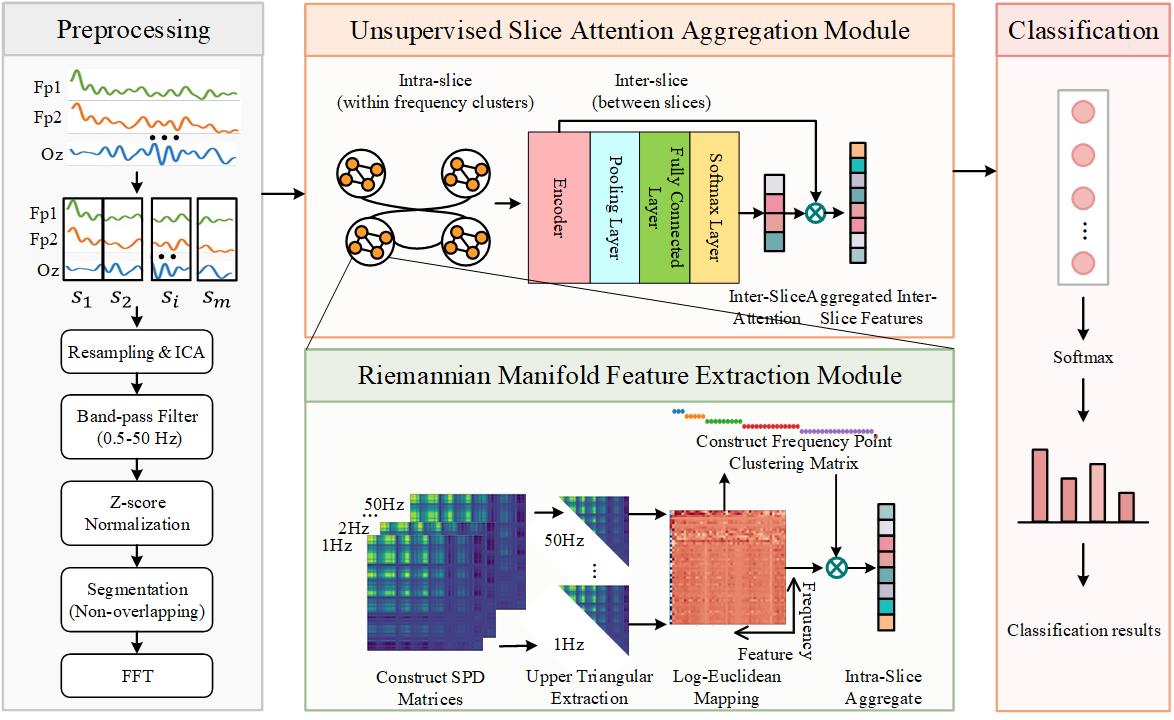}
    \caption{Overview of \method{}. The pipeline comprises four stages: Preprocessing and FFT; USAA Module to perform frequency cluster aggregation via K-Means and intra-inter slice attention fusion; RMFE Module to perform frequency-wise covariance construction and Log-Euclidean tangent-space mapping; Classification for mental stress detection.}
    \label{fig:framework}
\end{figure}

\method{} addresses cross-subject EEG stress detection through the two-module architecture illustrated in Figure~\ref{fig:framework}. Given raw multi-channel EEG signals of dimension $\R^{C \times L}$ ($C = 64$ channels), each trial is first segmented into $m$ non-overlapping slices and transformed into the frequency domain via FFT, yielding $\mathbf{X}_{\text{freq}} \in \R^{m \times C \times F}$ where $F$ is the number of frequency bins. Standard preprocessing steps including resampling, ICA artifact removal, bandpass filtering (0.5--50\,Hz), and z-score normalization are applied prior to segmentation.

The Riemannian Manifold Feature Extraction (RMFE) module constructs a covariance matrix $\mathbf{R}_{f} \in \SPD$ independently at each frequency point $f$, preserving frequency-specific spatial correlations that are critical for stress discrimination. Each SPD matrix is then mapped to the tangent space via the Log-Euclidean operator and vectorized as $\mathbf{h}_{f} \in \R^{D_0}$, yielding spectral-spatial features $\mathbf{H}_{\text{freq}} \in \R^{m \times F \times D_0}$ with $D_0 = C(C+1)/2$.

The Unsupervised Slice Attention Aggregation (USAA) module then processes these features in two stages. First, K-Means clustering groups correlated frequencies into $K$ clusters corresponding to canonical EEG bands; within-cluster weighted aggregation followed by across-cluster concatenation produces compact slice features $\mathbf{H} \in \R^{m \times D}$ with $D = K \times D_0$. Second, an intra-inter slice attention fusion mechanism adaptively weights the $m$ temporal slices: each slice feature is projected through a linear encoding layer to $\mathbf{H}_{\text{enc}} \in \R^{m \times d_{\text{enc}}}$, followed by channel-wise mean pooling and a fully connected softmax layer to compute slice-specific attention weights $\boldsymbol{\alpha}$. The weighted aggregation $\mathbf{H}_{\text{agg}} = \sum_{j} \alpha_j \mathbf{H}_{\text{enc}}(:, j, :)$ produces a fixed-dimensional trial-level representation, which is fed to a fully connected classifier for stress level prediction. All parameters are jointly optimized end-to-end via backpropagation.

\subsection{Frequency-Aware Riemannian Representation}

\paragraph{Motivation.} Traditional Riemannian EEG methods construct a single covariance matrix from time-domain signals, mixing frequency information across bands. For stress detection---a task where frequency-domain features have been shown to carry critical discriminative information~\cite{da2013eeg}---this obscures frequency-specific spatial patterns.

\paragraph{Frequency-wise covariance construction.} After FFT, the input tensor has dimension $\mathbf{X}_{\text{freq}} \in \R^{B \times m \times C \times F}$. For each frequency point $f \in \{1, \ldots, F\}$, we extract the spectral amplitude vector across all channels, as defined in Eq.~\ref{eq:freq_index}.
We then independently construct a covariance matrix from $\mathbf{x}_{b,j,:,f}$, adding $\varepsilon = 10^{-6}$ times the identity matrix $\mathbf{I}$ to ensure positive definiteness, as defined in Eq.~\ref{eq:freq_cov}.
This preserves frequency-specific spatial correlations that single time-domain covariance matrices would conflate.
\begin{equation}
    \mathbf{x}_{b,j,:,f} = \mathbf{X}_{\text{freq}}(b, j, :, f) \in \R^C
    \label{eq:freq_index}
\end{equation}
\begin{equation}
    \mathbf{R}_{b,j,f} = \mathbf{x}_{b,j,:,f} \, \mathbf{x}_{b,j,:,f}^\top + \varepsilon \mathbf{I}
    \label{eq:freq_cov}
\end{equation}

\paragraph{Log-Euclidean tangent-space mapping.} Each covariance matrix $\mathbf{R}_{b,j,f} \in \SPD$ resides on the SPD manifold, which lacks vector space structure. We employ the Log-Euclidean mapping to project onto the tangent space and extract the tangent-space feature vector, as defined in Eq.~\ref{eq:vech}:
\begin{equation}
    \mathbf{h}_{b,j,f} = \operatorname{vech}\!\left(\log(\mathbf{R}_{b,j,f})\right) \in \R^{D_0}
    \label{eq:vech}
\end{equation}
where $\log(\mathbf{R}_{b,j,f}) = \mathbf{U} \log(\boldsymbol{\Lambda}) \mathbf{U}^\top$ is computed via eigendecomposition $\mathbf{R}_{b,j,f} = \mathbf{U} \boldsymbol{\Lambda} \mathbf{U}^\top$ with $\mathbf{U} \in \R^{C \times C}$ and diagonal $\boldsymbol{\Lambda} \in \R^{C \times C}$, $\operatorname{vech}(\cdot)$ extracts the upper triangular portion, and $D_0 = C(C+1)/2 = 2080$ for $C = 64$. Collecting all frequency features yields $\mathbf{H}_{\text{freq}} \in \R^{B \times m \times F \times D_0}$, which serves as input to the frequency cluster aggregation module (Section~4.3).

\paragraph{Why this preserves spatial structure.} The covariance matrix encodes spatial information: diagonal elements reflect channel power, off-diagonal elements capture inter-channel functional connectivity~\cite{buzsaki2012origin}. The Log-Euclidean mapping preserves the geometric structure of the SPD manifold, whereas traditional vectorization in Euclidean space destroys this structure, leading to information loss that is amplified under cross-subject conditions.

\paragraph{Spatial-frequency duality.} A key insight from neuroscience is that low-frequency bands (delta, theta) reflect global brain-state modulation across widespread cortical regions, while high-frequency bands (beta, gamma) capture local, task-specific processing~\cite{buzsaki2012origin}. Under stress, high-frequency components exhibit greater sensitivity to state transitions~\cite{zhang2020stress}. This spatial-frequency duality---global structure in low frequencies, local structure in high frequencies---motivates our frequency-aware approach: by constructing covariance matrices independently per frequency point rather than collapsing across frequencies, we preserve both global and local spatial-frequency relationships that are critical for cross-subject stress detection.

\paragraph{Riemannian invariance for cross-subject robustness.} The Log-Euclidean mapping preserves the structure of the affine-invariant Riemannian metric on $\SPD$, which is invariant to affine transformations $\mathbf{R}_{b,j,f} \mapsto \mathbf{W}\mathbf{R}_{b,j,f}\mathbf{W}^\top$ for any invertible $\mathbf{W}$~\cite{barachant2011multiclass}. Consequently, our tangent-space features $\mathbf{h}_{b,j,f}$ (Eq.~\ref{eq:vech}) are inherently robust to inter-subject variations such as electrode displacement and impedance changes, and the subsequent cluster aggregation and attention fusion operate in a geometry-aware feature space where cross-subject variability is naturally mitigated.

\subsection{Frequency Cluster Aggregation}

\paragraph{Motivation.} Constructing $F$ covariance matrices yields high-dimensional features ($F \times D_0$), increasing computational cost. Moreover, individual frequency points carry limited independent physiological significance, as established EEG bands span multiple adjacent frequencies~\cite{da2013eeg}. We propose frequency cluster aggregation not as mere dimensionality reduction, but as a principled \textit{feature selection and redundancy elimination} mechanism that groups correlated frequency points into physiologically interpretable clusters.

\paragraph{Cluster determination Aggregation.} We apply the elbow method to determine the optimal number of clusters $K$ by analyzing the inertia curve across cluster counts. For MIST Control, $K = 5$; for MIST Stress, $K = 6$. In this section, $k \in \{1, \ldots, K\}$ denotes the cluster index and $K$ the total number of clusters.
We construct a frequency cluster assignment matrix $\mathbf{A} \in \R^{K \times F}$, where $A_{k,f}$ indicates the membership weight of frequency $f$ to cluster $k$. The aggregation proceeds in three steps: (1)~for each cluster $k$, the cluster feature is computed as the weighted sum of its member frequency features; (2)~the $K$ cluster features are concatenated along the feature dimension to form the slice representation $\mathbf{H}_{b,j}$; (3)~all $m$ slice representations are collected into the batch-level tensor $\mathbf{H}$, where $j \in \{1, \ldots, m\}$ indexes the temporal slices. The resulting $\mathbf{H}$ serves as input to the intra-inter slice fusion module (Section~4.4). The complete aggregation process is formalized as follows:
\begin{align}
    \mathbf{h}_{b,j,k} &= \sum_{f=1}^{F} A_{k,f} \cdot \mathbf{h}_{b,j,f} \label{eq:cluster_agg} \\
    \mathbf{H}_{b,j} &= [\mathbf{h}_{b,j,1}, \ldots, \mathbf{h}_{b,j,K}] \in \R^{D}, \quad D = K \times D_0 \label{eq:cluster_concat} \\
    \mathbf{H} &= [\mathbf{H}_{b,1}; \ldots; \mathbf{H}_{b,m}] \in \R^{B \times m \times D} \label{eq:slice_stack}
\end{align}

\paragraph{Physiological interpretability.} The data-driven clustering aligns with canonical EEG bands~\cite{britton2016electroencephalography}: Cluster~1 ($1$--$3$\,Hz) corresponds to delta, Cluster~2 ($4$--$12$\,Hz) spans theta and alpha, Cluster~3 ($13$--$25$\,Hz) maps to beta, Cluster~4 ($26$--$49$\,Hz) corresponds to gamma, and Cluster~5 ($50$\,Hz) captures line noise. This alignment validates that EEG frequency structure exhibits intrinsic cluster patterns, and our aggregation leverages this structure for stable, efficient representation. The geometric interpretation showing that frequency cluster aggregation performs principled data-driven grouping in Riemannian geometry is provided in Appendix~\ref{sec:geometric_interp}.

\subsection{Intra-Inter Slice Fusion}

\paragraph{Motivation.} After Riemannian feature extraction, each trial is represented as $m$ slice features $\mathbf{H} \in \R^{B \times m \times D}$ (Eq.~\ref{eq:slice_stack}). Existing methods aggregate slices via simple averaging or max pooling, assuming equal contribution from all slices. However, in stress detection, slices differ significantly in discriminative information---some capture critical state transitions while others contain mostly noise. Equal-weight aggregation dilutes informative signals and accumulates noise.

\paragraph{Slice number selection.} The slice number $m$ is a key hyperparameter that controls the granularity of temporal modeling. Increasing $m$ provides finer temporal resolution but incurs additional computation. Inspired by microeconomics---where the optimal consumption level is reached when marginal utility approaches zero---we define the marginal effect $\text{ME}(m)$ as the accuracy gain per unit of additional computation, where $\text{Acc}(m)$ denotes the accuracy with $m$ slices and $\text{FLOPs}(m)$ the corresponding computational cost, as defined in Eq.~\ref{eq:marginal}:
\begin{equation}
    \text{ME}(m) = \frac{\text{Acc}(m) - \text{Acc}(m-1)}{\text{FLOPs}(m) - \text{FLOPs}(m-1)}
    \label{eq:marginal}
\end{equation}
When $\text{ME}(m) > 0$, adding the $m$-th slice still improves accuracy; when $\text{ME}(m) \leq 0$, further slices yield no benefit. The optimal slice number $m^*$ is the largest $m$ with positive marginal effect, subject to a minimum accuracy threshold $\text{Acc}_{\min}$ and a computational budget $\text{FLOPs}_{\text{limit}}$, as defined in Eq.~\ref{eq:m_optimal}:
\begin{equation}
    m^* = \max\{m : \text{ME}(m) > 0\} \quad \text{s.t.} \quad \text{Acc}(m) \geq \text{Acc}_{\min}, \; \text{FLOPs}(m) \leq \text{FLOPs}_{\text{limit}}
    \label{eq:m_optimal}
\end{equation}

\paragraph{Intra-slice modeling.} When $m^* = 1$, the network processes a single slice, and the Riemannian feature extraction module alone provides intra-slice local spectral-spatial modeling. This corresponds to the R-\method{} variant in our ablation study.

\paragraph{Inter-slice attention fusion.} For $m^* > 1$, we propose an attention mechanism~\cite{vaswani2017attention} that adaptively weights $m^*$ temporal slices to highlight discriminative information. Let $d_{\text{enc}} = 128$ denote the encoding dimension. The attention fusion proceeds in four steps: (1)~linear encoding with weight $\mathbf{W} \in \R^{D \times d_{\text{enc}}}$ and bias $\mathbf{b} \in \R^{d_{\text{enc}}}$ projects slice features $\mathbf{H} \in \R^{B \times m^* \times D}$ to a lower-dimensional space; (2)~channel-wise mean pooling compresses each encoded slice to a scalar; (3)~a fully connected layer with softmax computes attention weights $\boldsymbol{\alpha} = [\alpha_1, \ldots, \alpha_{m^*}]^\top$ with $\sum_{j=1}^{m^*} \alpha_j = 1$; (4)~weighted aggregation produces the trial-level representation. Shared weights across the batch enhance generalization. The complete process is formalized as follows:
\begin{align}
    \mathbf{H}_{\text{enc}} &= \text{ReLU}(\mathbf{H} \mathbf{W} + \mathbf{b}) \in \R^{B \times m^* \times d_{\text{enc}}} \label{eq:linear_enc} \\
    \mathbf{h}_{\text{pool}}(:, j) &= \frac{1}{d_{\text{enc}}} \sum_{d=1}^{d_{\text{enc}}} \mathbf{H}_{\text{enc}}(:, j, d) \label{eq:pooling} \\
    \boldsymbol{\alpha} &= \text{Softmax}(\text{FC}(\mathbf{h}_{\text{pool}})) \in \R^{B \times m^*} \label{eq:attention} \\
    \mathbf{H}_{\text{agg}} &= \sum_{j=1}^{m^*} \alpha_j \cdot \mathbf{H}_{\text{enc}}(:, j, :) \in \R^{B \times d_{\text{enc}}} \label{eq:agg}
\end{align}
\subsection{Classification}
The classifier maps the aggregated features $\mathbf{H}_{\text{agg}} \in \R^{B \times d_{\text{enc}}}$ to stress level probabilities via a fully connected layer with weight $\mathbf{W}_c \in \R^{d_{\text{enc}} \times G}$ and bias $\mathbf{b}_c \in \R^G$, where $G$ is the number of stress classes. The model is trained by minimizing the cross-entropy loss over all $U$ trials, where $\mathbf{P}_u \in \R^G$ denotes the predicted probability vector for the $u$-th trial and $y_u \in \R^G$ is the one-hot label. Optimization uses Adam (learning rate $10^{-3}$, weight decay $10^{-2}$) for 300 epochs with batch normalization and dropout. The classification and loss are formalized as follows:
\begin{align}
    \mathbf{P} &= \text{Softmax}(\mathbf{W}_c \mathbf{H}_{\text{agg}} + \mathbf{b}_c) \in \R^{B \times G} \label{eq:classifier} \\
    \mathcal{L} &= -\sum_{u=1}^{U} y_u \log(\mathbf{P}_u) \label{eq:loss}
\end{align}

\section{Experimental Setup}

\noindent\textbf{Datasets.}
We evaluate on three datasets: \textbf{MIST Control} (30 subjects, 64 channels, 4-class stress levels), \textbf{MIST Stress} (same subjects under time-pressure and negative feedback, 4-class), and \textbf{SEED} (15 subjects, 62 channels, 3-class emotion). We adopt stratified 5-fold cross-validation at the subject level with zero subject overlap. Full dataset details and evaluation metrics are provided in Appendices~\ref{sec:dataset_details} and~\ref{sec:eval_metrics}.
\noindent\textbf{Baselines and Metrics.}
We compare against five baselines: EEGNet~\cite{lawhern2018eegnet}, BIOT~\cite{yang2023biot}, LaBraM~\cite{jiang2024large}, NeuroBOLT~\cite{li2024neurobolt}, and CorrAtt~\cite{hu2025correlation}. All models are trained from scratch under identical data splits. We adopt B.ACC, Precision, Recall, F1, and AUC as evaluation metrics. Baseline descriptions, metric definitions, and implementation details are provided in Appendices~\ref{sec:baseline_ref}--\ref{sec:impl_details}.


\subsection{Comparison with Existing Networks}

To demonstrate the effectiveness of \method{}, we compare it against five state-of-the-art (SOTA) baselines networks covering diverse architectures: EEGNet~\cite{lawhern2018eegnet} (compact CNN), BIOT~\cite{yang2023biot} (brain imaging transformer), LaBraM~\cite{jiang2024large} (large brain model), NeuroBOLT~\cite{li2024neurobolt} (bolt encoder), and CorrAtt~\cite{hu2025correlation} (correlation attention). All models are trained from scratch using identical data splits and training configurations.

Table~\ref{tab:main} summarizes the classification performance across three datasets. \method{} consistently achieves the best results across all metrics. On MIST Control, \method{} attains 77.59\% B.ACC with 95.61\% AUC, substantially outperforming the strongest baseline EEGNet (53.26\% B.ACC) by 24.33 percentage points and delivering 83.61\% precision and 75.79\% F1 score. On MIST Stress, which involves more subtle cognitive distinctions under induced pressure, \method{} reaches 75.88\% B.ACC and 94.57\% AUC, surpassing CorrAtt (46.67\% B.ACC) by 29.21\%---the largest margin among all datasets. On SEED, \method{} achieves 82.78\% B.ACC and 94.45\% AUC, exceeding EEGNet by 24.77\% while maintaining 83.74\% precision and 82.76\% F1.
The consistent improvements are particularly pronounced on MIST datasets, which employ cognitively complex paradigms with stronger low-frequency components~\cite{grissmann2017context}. In such scenarios, signal slicing aggravates information loss, making the preservation of spatial structure and temporal coherence critical---precisely what \method{} provides through frequency-aware Riemannian modeling and intra-inter slice fusion.

\begin{table*}[t]
\centering
\begin{minipage}[t]{0.4\textwidth}
    \centering
    \caption{Comparison of state-of-the-art Network on Three Datasets.}
    \label{tab:main}
    \scriptsize
    \setlength{\tabcolsep}{2pt}
    \renewcommand{\arraystretch}{0.95}
    \resizebox{\linewidth}{!}{%
    \begin{tabular}{lccccc}
        \toprule
        Method & B.ACC$\uparrow$ & Pre$\uparrow$ & F1$\uparrow$ & Rec$\uparrow$ & AUC$\uparrow$ \\
        \midrule
        \multicolumn{6}{c}{MIST Control} \\
        \midrule
        EEGNet$^{**}$ & 53.26$\pm$1.07 & 50.88$\pm$0.30 & 50.49$\pm$1.48 & 53.26$\pm$1.07 & 76.48$\pm$0.27 \\
        BIOT$^{**}$ & 37.09$\pm$4.63 & 40.99$\pm$1.04 & 37.00$\pm$4.13 & 38.37$\pm$4.17 & 67.03$\pm$0.85 \\
        LaBraM$^{**}$ & 34.08$\pm$2.59 & 35.08$\pm$3.18 & 34.08$\pm$3.93 & 34.15$\pm$3.71 & 59.14$\pm$3.62 \\
        NeuroBOLT$^{**}$ & 36.45$\pm$0.73 & 34.72$\pm$2.84 & 29.60$\pm$4.88 & 36.45$\pm$0.73 & 59.42$\pm$1.88 \\
        CorrAtt$^{**}$ & 45.37$\pm$1.75 & 42.02$\pm$1.62 & 35.37$\pm$2.36 & 39.86$\pm$1.92 & 68.23$\pm$1.77 \\
        \method{} & \textbf{77.59$\pm$2.31} & \textbf{83.61$\pm$2.77} & \textbf{75.79$\pm$3.33} & \textbf{74.11$\pm$2.70} & \textbf{95.61$\pm$0.96} \\
        \midrule
        \multicolumn{6}{c}{MIST Stress} \\
        \midrule
        EEGNet$^{**}$ & 48.82$\pm$1.21 & 50.57$\pm$1.55 & 48.50$\pm$1.47 & 52.28$\pm$1.28 & 72.89$\pm$1.61 \\
        BIOT$^{**}$ & 30.08$\pm$2.43 & 32.63$\pm$3.21 & 30.19$\pm$2.10 & 34.86$\pm$0.57 & 54.72$\pm$2.46 \\
        LaBraM$^{**}$ & 27.33$\pm$2.77 & 25.08$\pm$3.19 & 27.33$\pm$3.07 & 29.43$\pm$2.55 & 53.36$\pm$2.07 \\
        NeuroBOLT$^{**}$ & 29.56$\pm$0.64 & 31.52$\pm$2.93 & 29.20$\pm$1.98 & 33.67$\pm$0.65 & 51.25$\pm$1.73 \\
        CorrAtt$^{**}$ & 46.67$\pm$1.36 & 48.55$\pm$2.12 & 46.67$\pm$2.31 & 51.7$\pm$1.78 & 72.21$\pm$2.23 \\
        \method{} & \textbf{75.88$\pm$4.09} & \textbf{76.42$\pm$10.18} & \textbf{67.79$\pm$8.34} & \textbf{67.90$\pm$7.93} & \textbf{94.57$\pm$1.14} \\
        \midrule
        \multicolumn{6}{c}{SEED} \\
        \midrule
        EEGNet$^{**}$ & 58.01$\pm$1.75 & 58.48$\pm$3.25 & 57.22$\pm$1.30 & 53.01$\pm$1.75 & 75.20$\pm$2.09 \\
        BIOT$^{**}$ & 54.70$\pm$0.91 & 55.54$\pm$1.36 & 53.84$\pm$1.37 & 55.00$\pm$0.88 & 73.92$\pm$1.27 \\
        LaBraM$^{**}$ & 57.10$\pm$0.94 & 57.29$\pm$1.42 & 57.10$\pm$1.75 & 52.09$\pm$0.72 & 74.35$\pm$1.59 \\
        NeuroBOLT$^{**}$ & 50.23$\pm$0.99 & 52.94$\pm$3.37 & 49.76$\pm$1.57 & 50.23$\pm$0.99 & 69.28$\pm$1.32 \\
        CorrAtt$^{**}$ & 57.26$\pm$1.37 & 58.83$\pm$1.43 & 52.26$\pm$1.78 & 51.67$\pm$1.62 & 70.92$\pm$1.72 \\
        \method{} & \textbf{82.78$\pm$1.23} & \textbf{83.74$\pm$0.67} & \textbf{82.76$\pm$1.31} & \textbf{82.83$\pm$1.11} & \textbf{94.45$\pm$1.27} \\
        \bottomrule
    \end{tabular}%
    }
    \begin{tablenotes}[flushleft]
        \scriptsize
        \item Best results are highlighted in bold. $^{**}p < 0.01$ indicates significant accuracy differences between \method{} and SOTA.
    \end{tablenotes}
\end{minipage}
\hfill
\begin{minipage}[t]{0.59\textwidth}
    \centering
    \caption{Ablation Study Results on Three Datasets.}
    \label{tab:ablation}
    \scriptsize
    \setlength{\tabcolsep}{2pt}
    \renewcommand{\arraystretch}{0.95}
    \resizebox{\linewidth}{!}{%
    \begin{tabular}{cc l ccccc}
        \toprule
        R & I & Network & B.ACC & Pre & F1 & Rec & AUC \\
        \midrule
        \multicolumn{8}{c}{\textbf{MIST Control}} \\
        \midrule
        & & baseline$^{**}$ & 36.24$\pm$1.21 & 32.39$\pm$2.17 & 31.23$\pm$2.09 & 35.07$\pm$1.91 & 57.91$\pm$1.38 \\
        $\checkmark$ & & R-\method{}$^{**}$ & 44.47$\pm$0.78 & 44.78$\pm$0.90 & 42.66$\pm$0.85 & 42.11$\pm$0.85 & 69.51$\pm$0.78 \\
        & $\checkmark$ & I-\method{}$^{**}$ & 73.33$\pm$3.56 & 78.80$\pm$4.09 & 74.32$\pm$3.84 & 72.98$\pm$3.59 & 86.07$\pm$1.58 \\
        $\checkmark$ & $\checkmark$ & \method{} & \textbf{77.59$\pm$2.31} & \textbf{83.61$\pm$2.77} & \textbf{75.79$\pm$3.33} & \textbf{74.11$\pm$2.70} & \textbf{95.61$\pm$0.96} \\
        \midrule
        \multicolumn{8}{c}{\textbf{MIST Stress}} \\
        \midrule
        & & baseline$^{**}$ & 30.48$\pm$1.35 & 24.19$\pm$1.42 & 22.33$\pm$2.26 & 28.71$\pm$1.13 & 52.74$\pm$0.92 \\
        $\checkmark$ & & R-\method{}$^{**}$ & 40.43$\pm$0.44 & 42.60$\pm$2.31 & 37.77$\pm$0.40 & 36.80$\pm$0.26 & 62.81$\pm$0.83 \\
        & $\checkmark$ & I-\method{}$^{**}$ & 42.29$\pm$1.17 & 39.57$\pm$1.69 & 39.18$\pm$1.80 & 38.96$\pm$1.81 & 59.28$\pm$1.10 \\
        $\checkmark$ & $\checkmark$ & \method{} & \textbf{75.88$\pm$4.09} & \textbf{76.42$\pm$10.18} & \textbf{67.79$\pm$8.34} & \textbf{67.90$\pm$7.93} & \textbf{94.57$\pm$1.14} \\
        \midrule
        \multicolumn{8}{c}{\textbf{SEED}} \\
        \midrule
        & & baseline$^{**}$ & 39.13$\pm$0.34 & 39.22$\pm$0.31 & 39.08$\pm$0.36 & 39.21$\pm$0.33 & 55.99$\pm$0.30 \\
        $\checkmark$ & & R-\method{}$^{**}$ & 44.98$\pm$1.97 & 45.58$\pm$1.72 & 44.38$\pm$2.08 & 45.02$\pm$1.99 & 63.55$\pm$1.94 \\
        & $\checkmark$ & I-\method{}$^{**}$ & 58.84$\pm$4.19 & 61.19$\pm$3.85 & 57.68$\pm$4.85 & 58.69$\pm$4.14 & 72.63$\pm$1.54 \\
        $\checkmark$ & $\checkmark$ & \method{} & \textbf{82.78$\pm$1.23} & \textbf{83.74$\pm$0.67} & \textbf{82.76$\pm$1.31} & \textbf{82.83$\pm$1.11} & \textbf{94.45$\pm$1.27} \\
        \bottomrule
    \end{tabular}%
    }
    \begin{tablenotes}[flushleft]
        \scriptsize
        \item Best results are highlighted in bold. R: Riemannian module, I: Inter-slice fusion. $^{**}p < 0.01$ indicates significant accuracy differences between \method{} and other networks.
    \end{tablenotes}
\vskip-4ex
\end{minipage}
\end{table*}
\subsection{Ablation Study}
\label{sec:sensitivity}
To assess each component of \method{}, we design four ablation variants: \textbf{baseline} without Riemannian modeling or inter-slice fusion, \textbf{R-\method{}} using only the Riemannian module ($m{=}1$), \textbf{I-\method{}} using only inter-slice fusion, and the full \method{}. Results are reported in Table~\ref{tab:ablation}.
The Riemannian module consistently improves over baseline, increasing B.ACC by 8.23\%, 9.95\%, and 5.85\% on MIST Control, MIST Stress, and SEED, respectively. This shows that preserving the intrinsic covariance geometry of EEG channels enhances cross-subject generalization. Inter-slice fusion yields even larger gains, especially on MIST Control with a 37.09\% B.ACC improvement, highlighting the importance of modeling temporal coherence under fragmented slice-wise EEG information.
The two modules are also complementary. The full \method{} outperforms the stronger single-module variant, I-\method{}, by 4.26\%, 33.59\%, and 23.94\% B.ACC on the three datasets. This suggests that Riemannian features provide geometrically meaningful spatial representations that enable attention-based fusion to better exploit cross-slice dependencies. Overall, the full model achieves 77.59\%/75.88\%/82.78\% B.ACC and 95.61\%/94.57\%/94.45\% AUC, confirming the benefit of joint spatial-temporal modeling.

\subsection{Efficiency and Interpretability for Cross-Subject EEG Stress Detection}
Table~\ref{tab:efficiency} compares computational efficiency. \method{} requires only 1.60M parameters and 31.95M FLOPs ($m{=}20$), which is 11.9\% of EEGNet's FLOPs, 0.4\% of BIOT's, and 0.3\% of LaBraM's and NeuroBOLT's. Even with $m{=}1$, \method{} achieves 1.60M FLOPs while maintaining competitive accuracy. This demonstrates that geometry-preserving, attention-based feature integration is far more parameter-efficient than scaling up model capacity.

\begin{table}[t]
    \centering
    \begin{minipage}{0.45\linewidth}
        \centering
        \begin{threeparttable}
            \captionof{table}{Efficiency Comparison.}
            \label{tab:efficiency}
            \footnotesize
            \begin{tabular}{lcc}
                \toprule
                Model & Params (M)$\downarrow$ & FLOPs (M)$\downarrow$ \\
                \midrule
                EEGNet        & 0.05  & 268.10   \\
                BIOT          & 3.20  & 8140.23  \\
                LaBraM        & 6.23  & 11095.78 \\
                NeuroBOLT     & 10.45 & 10504.58 \\
                CorrAtt       & 0.22  & 1674.45  \\
                \method{} ($m{=}1$)  & 1.60  & 1.60     \\
                \method{} ($m{=}20$) & 1.60  & 31.95    \\
                \bottomrule
            \end{tabular}
            \begin{tablenotes}[flushleft]
                \footnotesize
                \item Params and FLOPs are reported in millions (M).
            \end{tablenotes}
        \end{threeparttable}
    \end{minipage}
    \hfill
    \begin{minipage}{0.52\linewidth}
        \centering
        \includegraphics[width=1\linewidth]{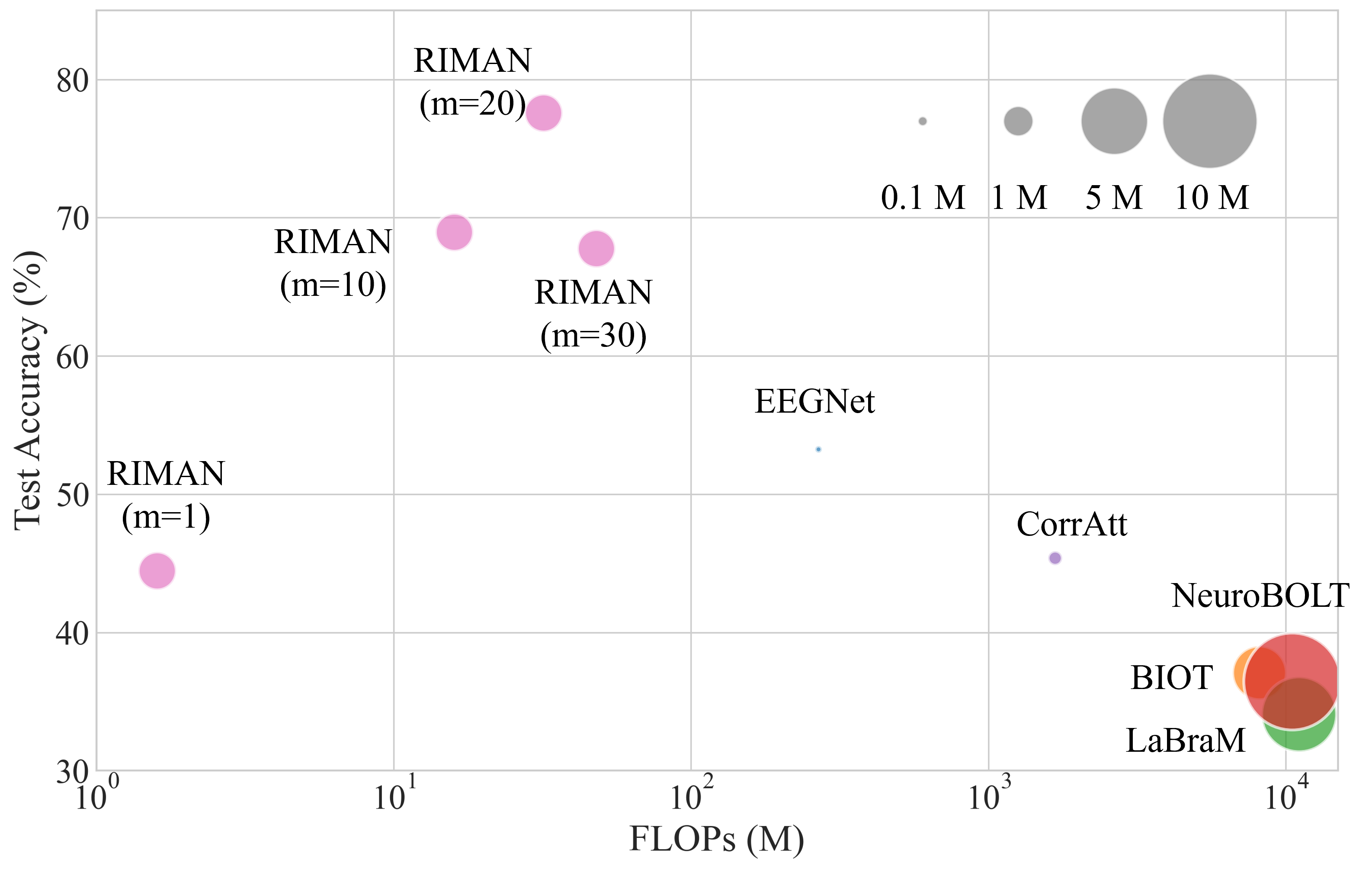}
        \captionof{figure}{Model Performance Comparison)}
        \label{fig:fig6}
    \end{minipage}
    \vskip-4ex
\end{table}

To further visualize the performance--efficiency trade-off, Figure~\ref{fig:fig6} presents a bubble chart where the x-axis (log scale) denotes FLOPs, the y-axis represents test accuracy on MIST Control, and bubble size encodes parameter count. \method{} (magenta bubbles) occupies the upper-left region---the Pareto-optimal zone of low computational cost and high accuracy. Specifically, \method{} with $m{=}20$ achieves 77.59\% accuracy at merely 31.95M FLOPs and 1.60M parameters, which amounts to only 11.9\% of EEGNet's FLOPs (268.10M) and less than 0.4\% of large-scale baselines such as BIOT (8140.23M), LaBraM (11095.78M), and NeuroBOLT (10504.58M). Even as $m$ increases from 1 to 30, \method{} variants remain tightly clustered in the low-FLOPs regime ($<$50M) while monotonically improving accuracy---a scalability property that baseline models, which reside in the high-FLOPs / lower-accuracy quadrant, cannot match. This geometry-preserving design thus delivers order-of-magnitude efficiency gains over brute-force capacity scaling, making \method{} particularly attractive for resource-constrained deployment scenarios such as wearable brain-computer interfaces.

\subsection{Channel Importance Topography Maps.}
\begin{figure}[htbp]
    \centering
    \begin{subfigure}{0.32\linewidth}  
        \centering
        \includegraphics[width=\linewidth]{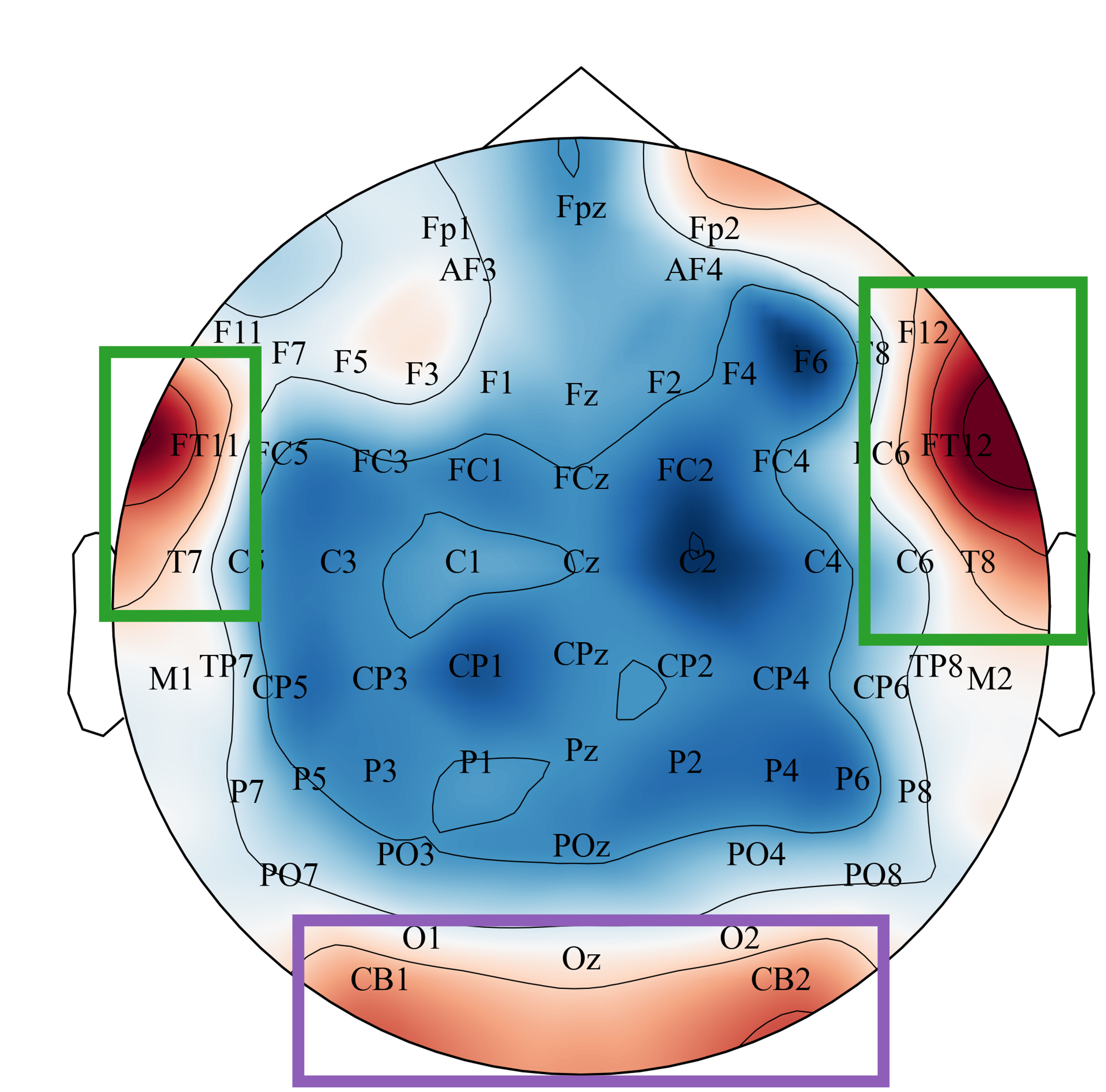}
        \caption{MIST Control Datasets}  
    \end{subfigure}
    \hfill  
    \begin{subfigure}{0.32\linewidth}
        \centering
        \includegraphics[width=\linewidth]{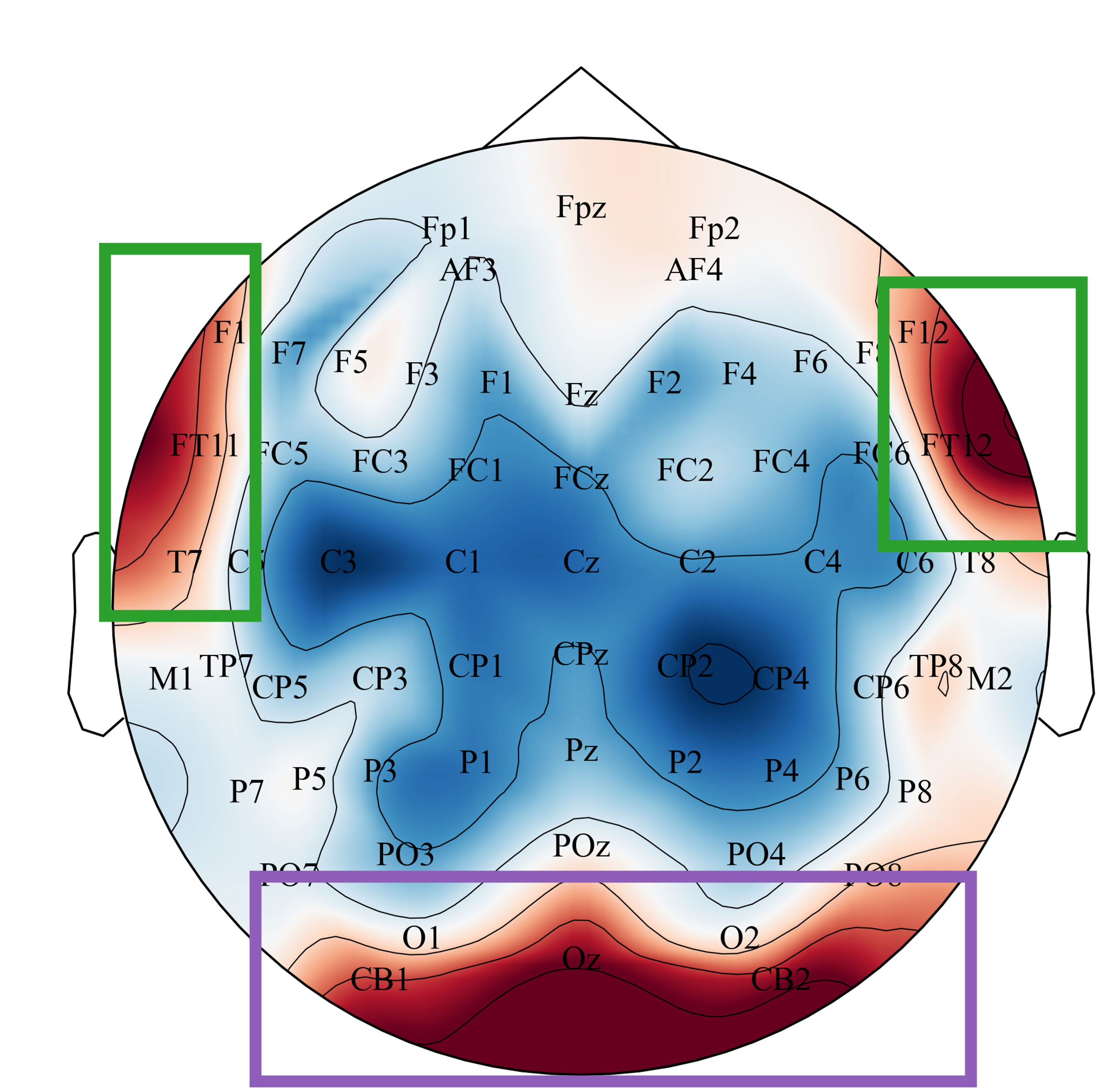}
        \caption{MIST Stress Datasets}  
    \end{subfigure}
    \hfill
    \begin{subfigure}{0.32\linewidth}
        \centering
        \includegraphics[width=\linewidth]{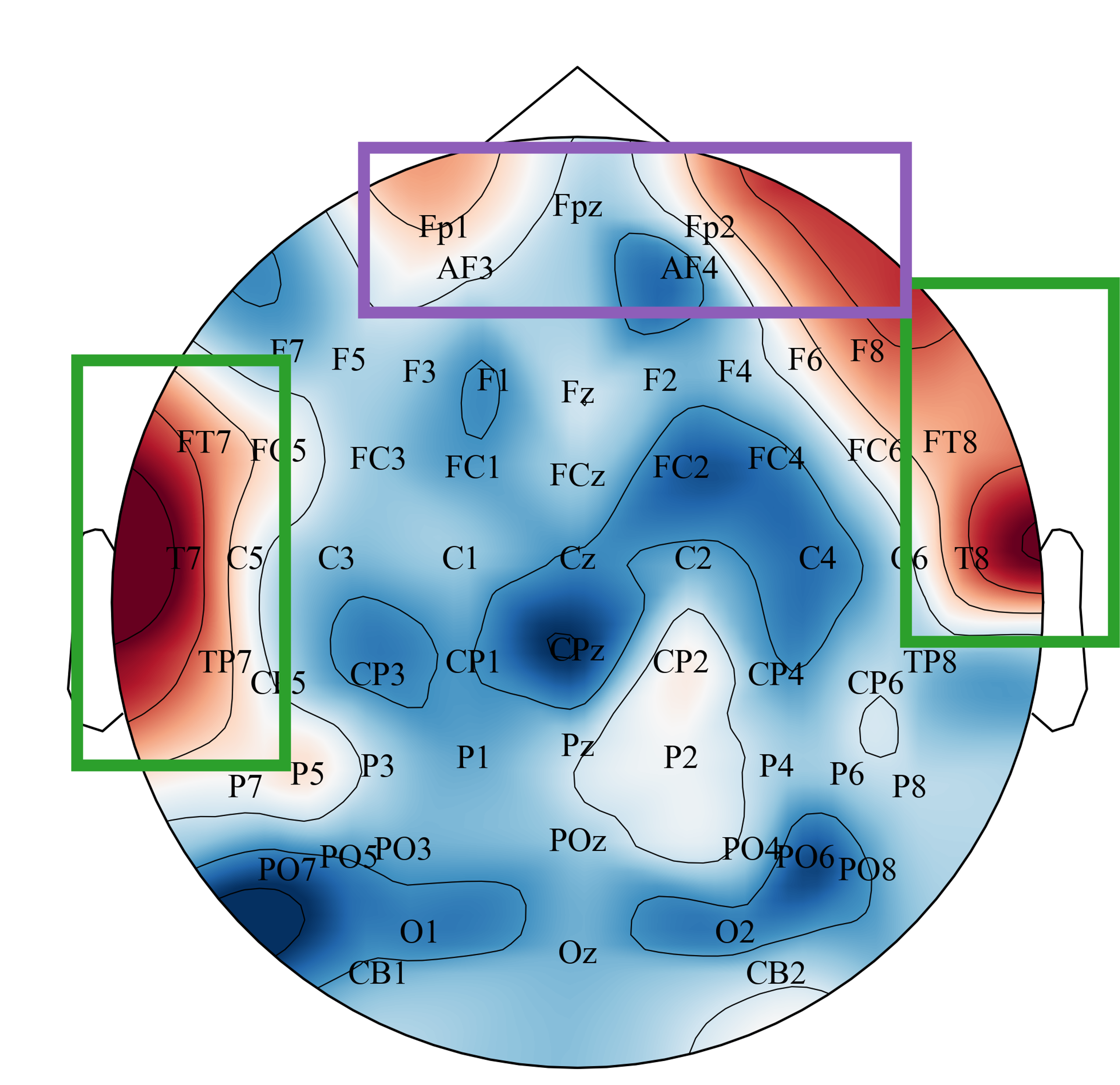}
        \caption{SEED Datasets}  
    \end{subfigure}
    \caption{Channel Importance Topographic Maps of \method{} in MIST Control, MIST Stress and SEED Datasets}
    \label{fig:fig7}
\end{figure}

To interpret which EEG channels \method{} relies on for classification, we visualize the learned channel attention weights as scalp topographic maps in Figure~\ref{fig:fig7}. The revealed attention patterns align closely with the cognitive demands of each paradigm.

On MIST Control (Figure~\ref{fig:fig7}a), \method{} assigns the highest importance to bilateral frontotemporal channels (FT11, FT12; green boxes) and occipitocerebellar channels (CB1, CB2; purple boxes). The frontotemporal emphasis reflects their established role in cognitive control and executive function during sustained mental arithmetic~\cite{davidson2000emotion}, while the occipital/cerebellar activation is consistent with visual attention maintenance and sensorimotor coordination required by the MIST task.

The MIST Stress condition (Figure~\ref{fig:fig7}b) exhibits a distinct reconfiguration: while frontotemporal channels (FT11, FT12; green boxes) remain highly weighted---now subserving stress regulation in addition to cognitive control---the occipital focus shifts centrally to Oz (purple box), and overall occipital sensitivity intensifies. This pattern is consistent with enhanced visual processing of evaluative feedback (e.g., error messages and time-pressure warnings) that characterizes the stress induction mechanism~\cite{ulrich2009neural}, suggesting that \method{} adaptively recruits visual cortical resources when emotional salience increases.
In contrast, SEED (Figure~\ref{fig:fig7}c) engages a qualitatively different network: temporal channels (T7, T8; green boxes) and prefrontal channels (Fp1, Fp2; purple boxes) dominate the attention map. This topology aligns with canonical emotion processing circuits~\cite{davidson2000emotion}, where the temporal lobes support emotional memory and semantic appraisal, and the prefrontal cortex regulates affective responses---a pattern markedly different from the cognitively-driven MIST topographies.


\section{Discussion and Limitation}

\paragraph{Why does Riemannian representation benefit cross-subject generalization?}
The affine-invariant Riemannian metric on the SPD manifold possesses inherent invariance to affine transformations~\cite{barachant2011multiclass}, providing robustness to inter-subject variations such as electrode displacement. Our ablation confirms that even the Riemannian-only variant (R-\method{}) improves over baseline by 8--10 percentage points, demonstrating that spatial structure preservation is critical for cross-subject transfer. The frequency-wise construction preserves spatial-frequency duality: under stress, high-frequency components (beta, gamma) exhibit greater sensitivity to state transitions~\cite{zhang2020stress}, enabling capture of stress-specific local cortical patterns.

\paragraph{Why does intra-inter slice fusion outperform single-slice modeling?}
Stress formation follows a dynamic allostatic process~\cite{juster2010allostatic} where different temporal windows contribute unequally. Intra-slice modeling captures local spectral patterns, while inter-slice modeling captures global temporal coherence~\cite{he2026dicha}. The ablation shows that I-\method{} yields 37.09\% improvement on MIST Control, and the full model further improves by 4.26\%, confirming complementarity. Moderate increases in slice count stabilize low-frequency feature estimation and capture cross-cycle temporal coherence~\cite{wu2025analysing}.

\paragraph{Limitations.} (1)~Our evaluation covers stress detection and emotion classification; generalization to other EEG decoding tasks (motor imagery, seizure detection) requires further validation. (2)~Dataset sizes remain limited (30 and 15 subjects); larger-scale studies are needed.

\section{Conclusion}

We propose \method{}, an Intra-Inter Riemannian Manifold Attention Network for cross-subject EEG stress detection. EEG signal segmentation into temporal slices inevitably causes information loss and insufficient feature extraction. \method{} addresses this challenge via two complementary mechanisms: spatial structure preservation via frequency-aware Riemannian modeling with cluster-based feature selection, and temporal coherence recovery via intra-inter slice attention fusion. On three datasets, \method{} achieves SOTA performance (B.ACC: 77.59\%/75.88\%/82.78\%) with only 1.60M parameters and 31.95M FLOPs, demonstrating that geometry-preserving, attention-based feature integration is both effective and efficient. \method{} provides a detection foundation for closed-loop stress regulation systems, with future work targeting online deployment and multi-modal fusion.

\section*{ACKNOWLEDGMENTS}
This work was supported in part by the Beijing Natural Science Foundation under Grant 4242033 and partially funded by the Deutsche Forschungsgemeinschaft (DFG, German Research Foundation) – SFB 1574 – 471687386.


\bibliography{references}
\bibliographystyle{plain}

\section*{Appendix}
\appendix

\section{Overview}
This appendix provides supplementary material organized as follows: Section~\ref{sec:dataset_details} describes the datasets and cross-subject protocol; Section~\ref{sec:eval_metrics} defines evaluation metrics; Section~\ref{sec:baseline_ref} details the baseline methods; Section~\ref{sec:impl_details} provides implementation details; Section~\ref{sec:geometric_interp} presents the geometric interpretation of frequency cluster aggregation; Sections~\ref{sec:freq_discriminability}--\ref{sec:sensitivity} contain additional spectral and sensitivity analyses; and Section~\ref{sec:broader_impact} discusses broader societal impacts.

\section{Dataset Details}
\label{sec:dataset_details}

\noindent\textbf{MIST Dataset.} This dataset contains EEG recordings from 30 healthy subjects performing mental arithmetic tasks under two conditions. EEG signals were recorded using 64 channels at 1000\,Hz (downsampled to 200\,Hz). Each session consists of a 20-minute mental arithmetic task subdivided into four difficulty levels. The two sessions differ in experimental manipulation: \textbf{MIST Control} imposes no time constraints or evaluative feedback, while \textbf{MIST Stress} introduces strict time limits and negative social-evaluative feedback. The two sessions are separated by $\geq 7$ days to minimize learning effects. We treat them as independent datasets due to the systematic differences in cognitive load and emotional stress. Both use 4-class stress labeling based on difficulty levels.

\noindent\textbf{SEED Dataset.} This dataset comprises EEG recordings from 15 subjects (8 females, 7 males; mean age $23.27 \pm 2.37$ years) during an emotion-eliciting video-watching task. EEG signals were recorded from 62 electrodes at 1000\,Hz and downsampled to 200\,Hz. Each subject watched 15 film clips (5 positive, 5 neutral, 5 negative), with 3-class emotion labels.

\noindent\textbf{Cross-subject protocol.} We adopt stratified 5-fold cross-validation (CV) at the subject level: subjects are partitioned into 5 disjoint folds preserving label proportions. In each fold, 4 folds serve as training subjects and 1 fold as test subjects, with zero subject overlap. Data is split 8:2 into training and test sets; the training set undergoes 5-fold CV for model selection, followed by retraining on the full training set.

\section{Evaluation Metrics}
\label{sec:eval_metrics}

We adopt five metrics: \textbf{Balanced Accuracy (B.ACC)} averages per-class recall to handle class imbalance, defined as $\text{B.ACC} = \frac{1}{G}\sum_{g=1}^{G} \text{Rec}_g$; \textbf{Precision (Pre)} measures the fraction of correct positive predictions; \textbf{Recall (Rec)} measures the fraction of actual positives correctly identified; \textbf{F1-Score (F1)} is the harmonic mean of precision and recall; and \textbf{AUC} computes the area under the receiver operating characteristic curve. All metrics are reported as percentages.

\section{Baseline Reference}
\label{sec:baseline_ref}

We compare against five representative methods: \textbf{EEGNet}~\cite{lawhern2018eegnet} is a compact CNN with depthwise and separable convolutions (0.05M params); \textbf{BIOT}~\cite{yang2023biot} is a biosignal transformer with channel embedding alignment for cross-subject generalization (3.20M params); \textbf{LaBraM}~\cite{jiang2024large} is a large-scale pretrained brain model with masked EEG modeling (6.23M params); \textbf{NeuroBOLT}~\cite{li2024neurobolt} synthesizes fMRI signals from raw EEG via multi-dimensional representation learning (10.45M params); and \textbf{CorrAtt}~\cite{hu2025correlation} employs correlation-matrix self-attention for EEG classification (0.22M params). All models are trained from scratch under identical data splits and training configurations.

\section{Implementation Details}
\label{sec:impl_details}

\textbf{Preprocessing.} Resampling to 200\,Hz, ICA artifact removal, 8th-order Butterworth bandpass filter (0.5--50\,Hz), z-score normalization, 8-second non-overlapping segmentation, and covariance computation. The slice number $m$ is set to 29 for MIST Control, 28 for MIST Stress, and 26 for SEED, determined via marginal effect optimization (Eq.~\ref{eq:m_optimal}).

\textbf{Training.} The model is optimized with Adam (learning rate $10^{-3}$, weight decay $10^{-2}$) for 300 epochs with batch normalization and dropout.

\textbf{Hardware and Software.} AMD Ryzen 9 3950X CPU, NVIDIA RTX 6000 GPU, 64\,GB RAM. Software: Ubuntu 20.04, Python 3.8, PyTorch 1.12, MNE-Python 0.24.

\section{Geometric Interpretation of Frequency Cluster Aggregation}
\label{sec:geometric_interp}

Since the tangent-space features $\mathbf{h}_{b,j,f}$ are obtained via the Log-Euclidean map from the SPD manifold, the Euclidean distance between any two tangent-space vectors $\mathbf{h}_{b,j,f_1}, \mathbf{h}_{b,j,f_2}$ equals the Log-Euclidean distance between their corresponding SPD matrices on the manifold, i.e., $\|\mathbf{h}_{b,j,f_1} - \mathbf{h}_{b,j,f_2}\|_2 = d_{\text{LE}}(\mathbf{R}_{b,j,f_1}, \mathbf{R}_{b,j,f_2})$. Consequently, minimizing the K-Means objective in tangent space is equivalent to grouping frequency points according to their Riemannian distance on $\SPD$. Moreover, the cluster feature $\mathbf{h}_{b,j,k}$ (Eq.~\ref{eq:cluster_agg}) is exactly the tangent-space representation of the Log-Euclidean Fr\'{e}chet mean of cluster $k$ on $\SPD$. Define the Fr\'{e}chet mean as $\bar{\mathbf{R}}_k = \exp(\sum_{f} A_{k,f} \cdot \log(\mathbf{R}_{b,j,f})) \in \SPD$; then its tangent-space feature satisfies:
\begin{equation}
    \operatorname{vech}\!\left(\log(\bar{\mathbf{R}}_k)\right) = \sum_{f=1}^{F} A_{k,f} \cdot \mathbf{h}_{b,j,f} = \mathbf{h}_{b,j,k}
    \label{eq:frechet_exact}
\end{equation}
where the first equality follows from $\log(\exp(\cdot)) = \cdot$ and the linearity of $\operatorname{vech}(\cdot)$, and the second from Eq.~\ref{eq:cluster_agg}. This establishes that frequency cluster aggregation performs principled data-driven grouping in Riemannian geometry rather than arbitrary dimensionality reduction, with each cluster feature precisely corresponding to the tangent-space representation of the Log-Euclidean Fr\'{e}chet mean on $\SPD$.

\section{Frequency-based Discriminability and Temporal Dynamics}
\label{sec:freq_discriminability}
This section investigates the impact of temporal window length on the stability of frequency-domain representations and the discriminative power of EEG features. Analysis is conducted across all datasets, each of which is sampled at 200 Hz. Window lengths vary from 100 to 3200 samples (0.5 s to 16 s), spanning durations from brief segments that capture fast transient dynamics to windows that encompass two full 8 s for MIST Stress arithmetic cycles—ensuring both robust estimation of low-frequency activity and assessment of inter-cycle coherence.

\begin{figure*}[t!]
    \centering
    \includegraphics[width=0.8\linewidth]{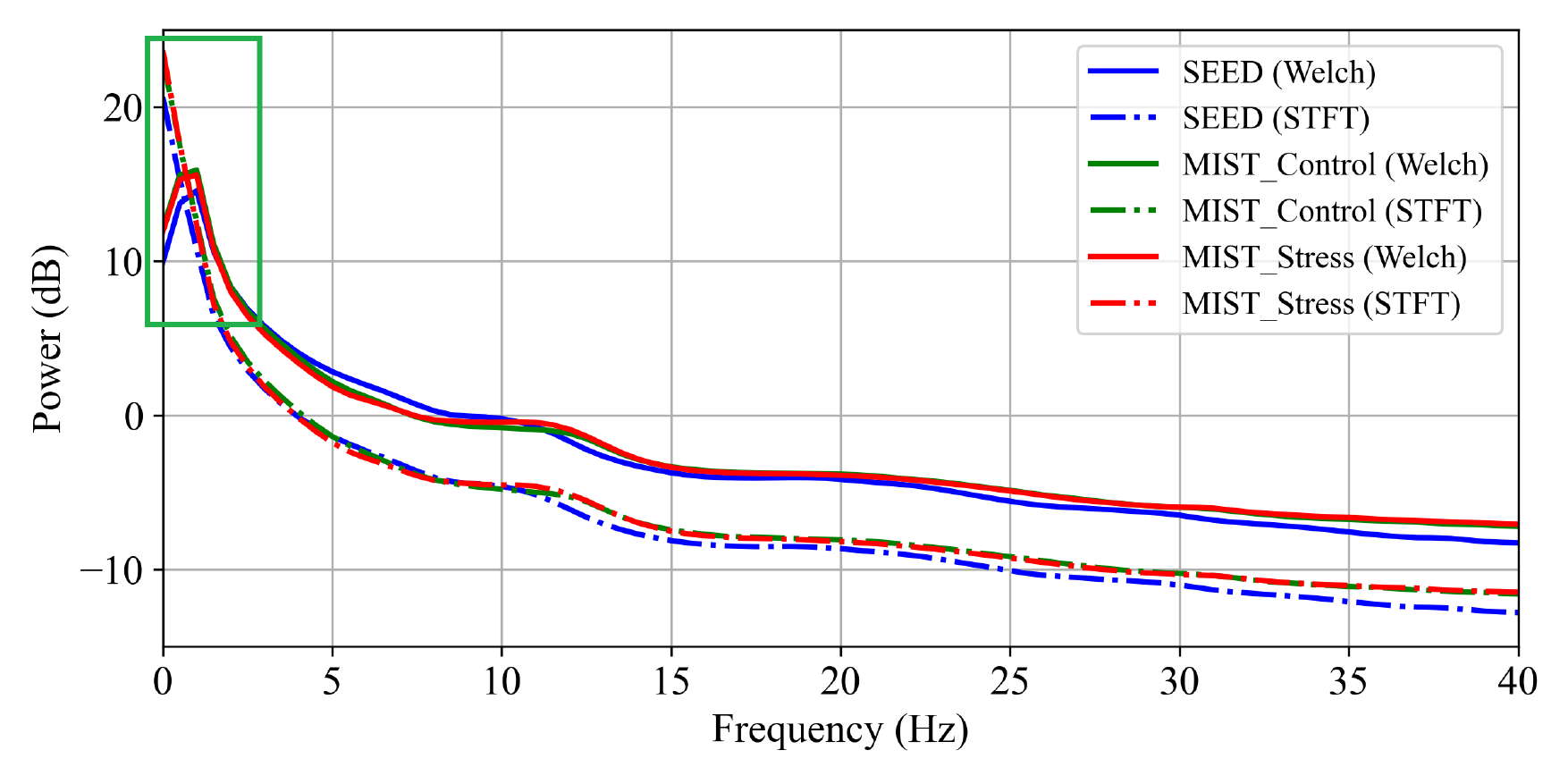}
    \caption{Topographical Maps of Channel Discriminability Across Temporal Windows (Welch Spectral Features).}
    \label{fig:fig2}
\end{figure*}

\paragraph{ Effect of Slice Length on Frequency Information Preservation.}
Spectral analysis in Figure~\ref{fig:fig2} (x-axis: frequency in Hz; y-axis: power in dB) reveals that EEG signals are dominated by low-frequency components. The green box highlights regions with strong low-frequency power, which are more susceptible to being distorted by short slices. Welch's method and STFT are employed to effectively capture frequency information under different temporal conditions. The former provides stable spectral estimates by averaging periodograms, while STFT retains temporal dynamics in non-stationary signals through a time-frequency representation. These methods complement each other, ensuring robust feature extraction across varying slice lengths. Short slice windows fail to retain low-frequency energy (e.g., unstable spectral peaks), whereas \method{}'s intra-/interslice modeling preserves both local dynamics and global coherence across varying temporal granularities. This enables superior performance over BIOT on MIST/SEED datasets, addressing slicing-induced temporal inconsistencies.

\section{Spectral Discriminability Across various Channels.}
\begin{figure}[t!]
    \centering
    \includegraphics[width=1.0\linewidth]{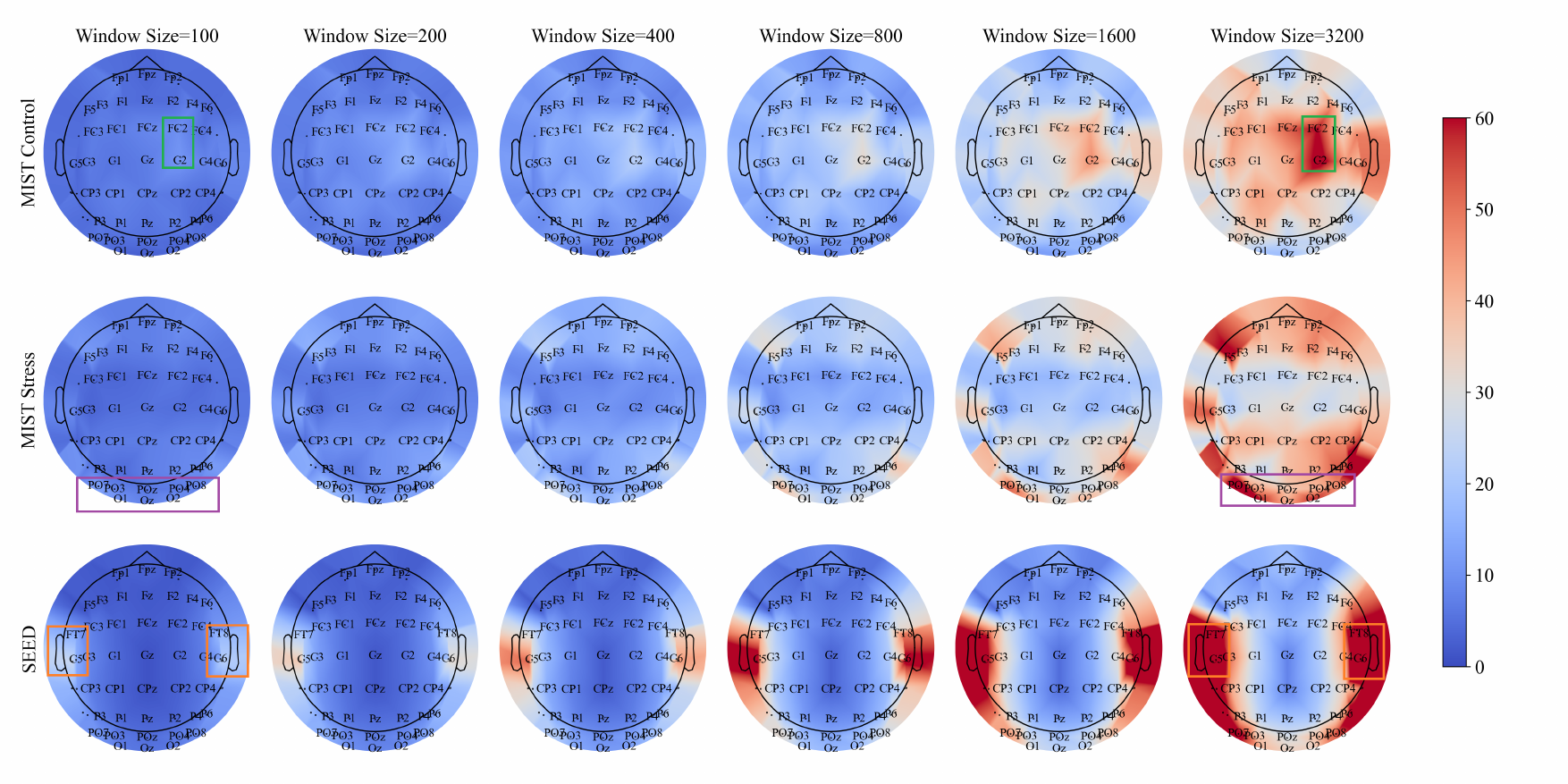}
    \caption{Comparison across Datasets and Methods in Spectrum domain.}
    \label{fig:fig3}
\end{figure}

Figure~\ref{fig:fig3} visualizes the class-wise discriminability of EEG channels under different window lengths using Welch power spectral features. For each window size, we compute the average Euclidean distance between classes for each channel. Warmer colors (red) indicate higher discriminability, while cooler colors (blue) denote lower relevance.
Green boxes highlight FC2 and C2 in the MIST Control paradigm, where extending the window from 100 to 3200 samples markedly increases inter-class distance in frontal electrodes linked to mental arithmetic load.  

Purple boxes mark O1 and O2 in the MIST Stress paradigm, suggseting similar gains in occipital channels that likely reflect enhanced visual processing of error/time-limit feedback under stress. Orange boxes surround C5, C6, FT7 and FT8 in the SEED dataset, where longer windows length boosts separability in temporal electrodes associated with emotion processing.

Across all datasets, larger window lengths yield richer low-frequency estimates and amplify region-specific discriminative patterns. This demonstrates that appropriately extending slice duration is critical for early detection of mental stress and affective states, as it stabilizes low-frequency features and captures cross-cycle coherence in channels.

\begin{figure}[t!]
    \centering
    \includegraphics[width=0.8\linewidth]{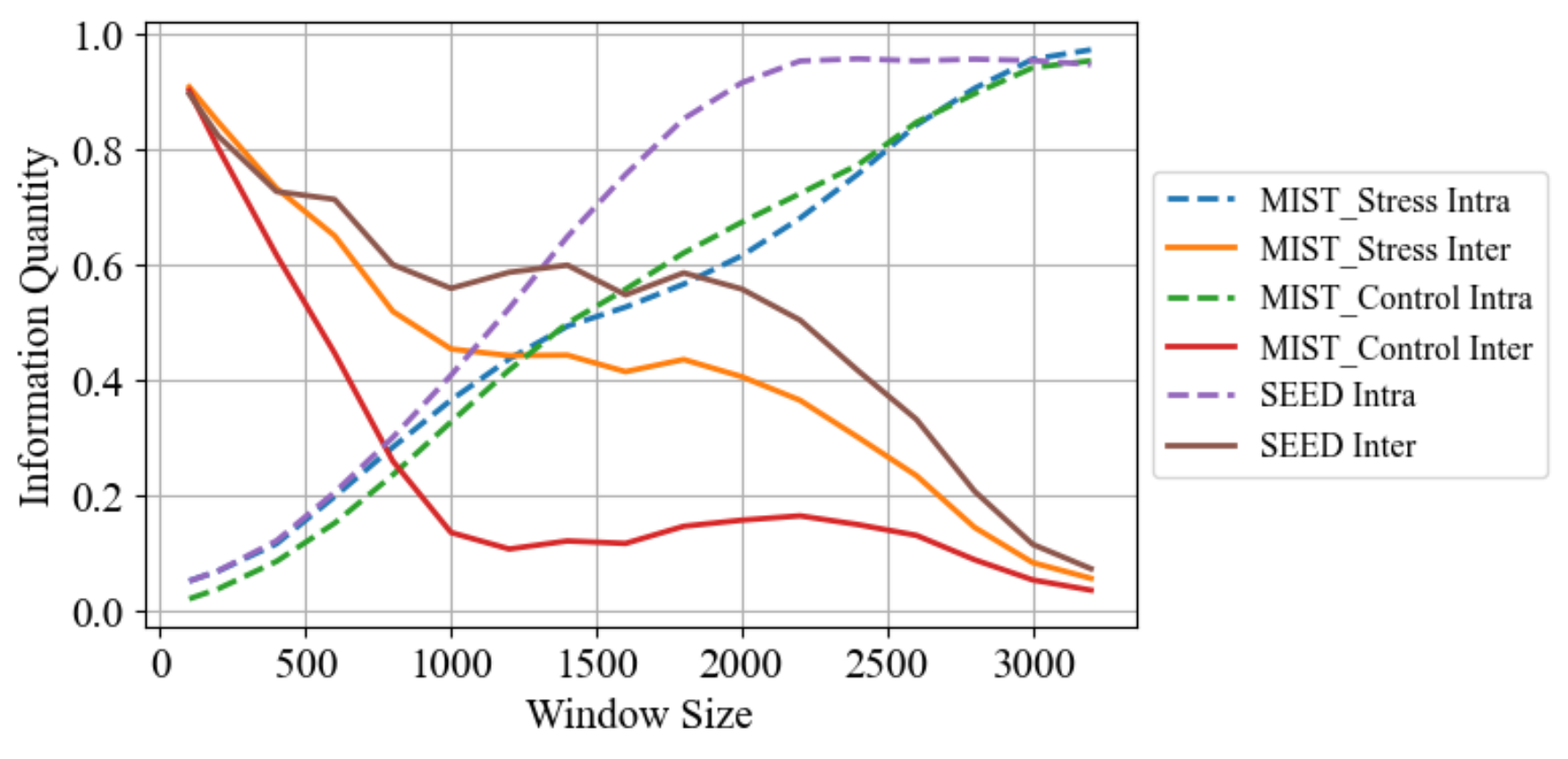}
    \caption{Comparison across Datasets and Methods in Spectrum domain.}
    \label{fig:fig4}
\end{figure}

\paragraph{Intraslice and Interslice Information.}
To further quantify the effect of window length on temporal stability and spectral dynamics (as visualized in Figure~\ref{fig:fig4}), we introduce two metrics: \emph{Intraslice Information} and \emph{Interslice Information}.

Intraslice Information measures the Euclidean distance between average spectral representations of different classes within the same temporal slice, reflecting inter-class separability and intra-class consistency; higher values indicate more distinguishable class-specific patterns.

Interslice Information captures local temporal dynamics by computing the mean spectral distance between adjacent slices of the same class; smaller values suggest more stable temporal evolution, while larger values indicate higher dynamism. Formally, let $\bar{\mathbf{h}}_{c,j} = \frac{1}{N_c} \sum_{i \in \mathcal{S}_c} \mathbf{h}_{i,j}$ denote the mean spectral representation of class $c$ in slice $j$, where $N_c$ is the number of samples of class $c$, $C$ is the number of classes and $m$ is the number of slices. The two metrics are defined as:
\begin{align}
    I_{\text{intra}}(j) &= \frac{1}{C(C-1)} \sum_{c_1=1}^{C} \sum_{c_2 > c_1}^{C} \|\bar{\mathbf{h}}_{c_1,j} - \bar{\mathbf{h}}_{c_2,j}\|_2 \label{eq:intraslice} \\
    I_{\text{inter}}(c) &= \frac{1}{m-1} \sum_{j=1}^{m-1} \|\bar{\mathbf{h}}_{c,j} - \bar{\mathbf{h}}_{c,j+1}\|_2 \label{eq:interslice}
\end{align}

As window size increases, Intraslice Information increases and Interslice Information decreases across all datasets, indicating that longer windows enhance class separability.

\section{Sensitivity to slice number.}
\label{sec:sensitivity}
\begin{figure}
    \centering
    \includegraphics[width=0.8\linewidth]{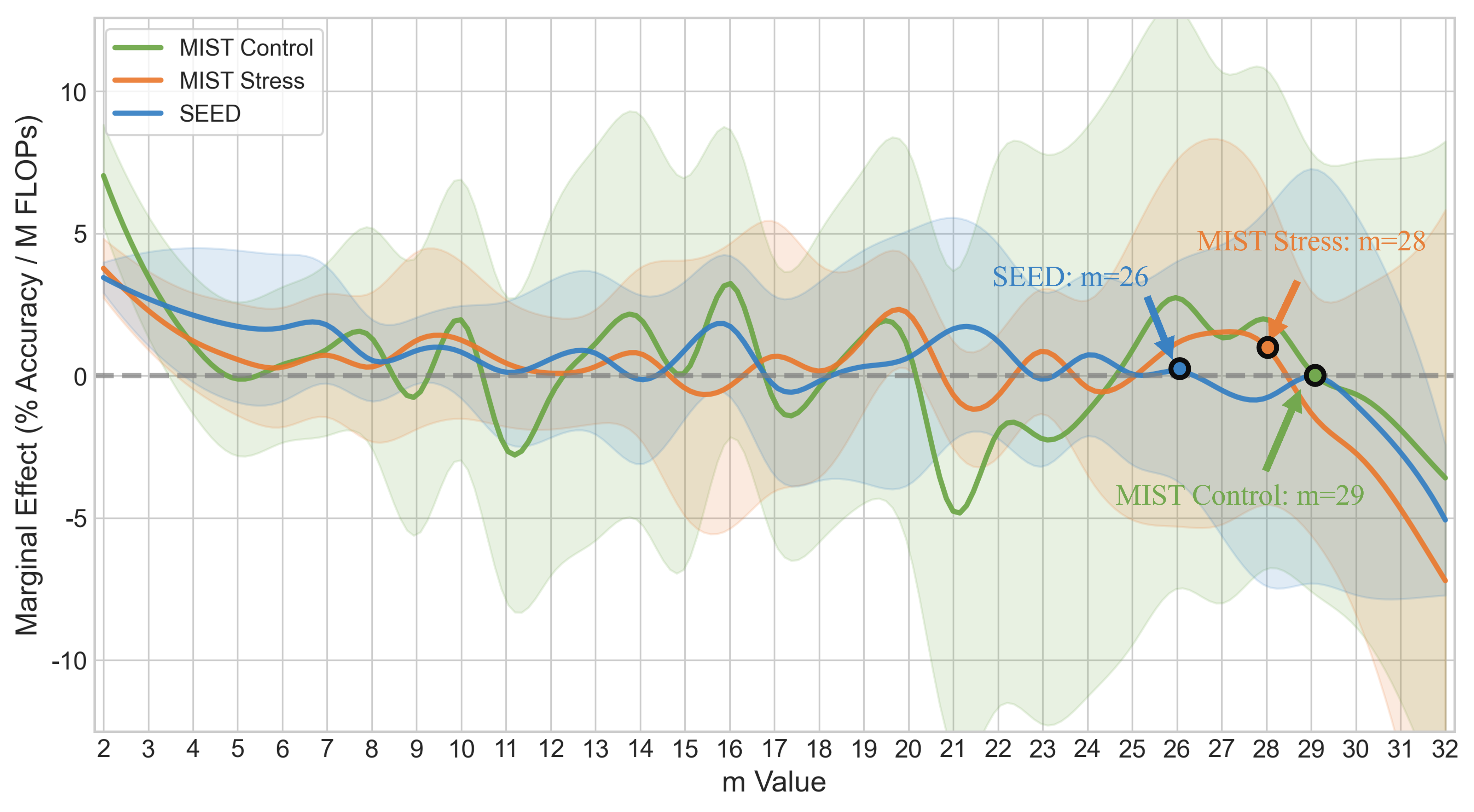}
    \caption{Marginal Effect Comparison across Three Datasets}
    \label{fig:fig5}
\end{figure}

The slice number $m$ is a key hyperparameter controlling the granularity of temporal modeling. We sweep $m \in \{1, \ldots, 32\}$ and quantify the computational efficiency of each setting via the marginal effect criterion defined in Eq.~\ref{eq:marginal}, which measures accuracy gain per unit of additional FLOPs. Figure~\ref{fig:fig5} plots the marginal effect curves for all three datasets, where the shaded regions denote variance across cross-validation folds. All three curves exhibit a consistent pattern: marginal effect rises initially as additional slices provide richer temporal information, peaks at an intermediate range, then declines---and eventually turns negative---as redundant slices introduce noise without commensurate accuracy gains.

Applying the constrained optimization formulation in Eq.~\ref{eq:m_optimal}, which maximizes marginal effect subject to minimum accuracy requirements, yields $m^* = 29$ for MIST Control, $m^* = 28$ for MIST Stress, and $m^* = 26$ for SEED, as annotated in Figure~\ref{fig:fig5}. These optimal points reside near where each curve crosses zero from above, indicating the sweet spot before further slices degrade cost-effectiveness. Notably, MIST datasets tolerate larger $m$ values than SEED, reflecting their longer task durations and richer temporal dynamics that benefit from finer slicing.



\section{Broader Impact}
\label{sec:broader_impact}

This work has both positive and negative societal implications. On the positive side, \method{} enables objective, portable mental stress monitoring that could benefit early intervention for stress-related disorders and support closed-loop stress regulation systems, particularly in resource-constrained settings where its low computational cost (1.60M parameters) enables deployment on wearable devices. On the negative side, EEG-based stress detection could be misused for workplace surveillance or discriminatory screening without informed consent. Cross-subject generalization limitations may also lead to biased assessments for underrepresented populations not well-represented in training data. We strongly advocate that any real-world deployment of \method{} requires informed consent, regulatory oversight, and fairness auditing across demographic groups.

\newpage

\end{document}